\documentclass[letterpaper, 10 pt, conference]{ieeeconf}

\usepackage{cite}
\usepackage[pdftex]{graphicx,graphics} 
\DeclareGraphicsExtensions{.pdf,.jpeg,.png}

\graphicspath{{figures/}}

\usepackage[cmex10]{amsmath}
\usepackage{amssymb}
\usepackage{mathtools}
\mathtoolsset{showonlyrefs=true}

\usepackage{array}
\usepackage{multirow}

\usepackage[caption=false,font=footnotesize]{subfig}
\usepackage{float}

\usepackage{url}

\usepackage{xcolor}
\usepackage{multirow}
\usepackage{color, colortbl}

\definecolor{Gray}{gray}{0.9}
\definecolor{White}{rgb}{1.0, 1.0, 1.0}
\definecolor{LightGray}{gray}{0.8}
\definecolor{LightCyan}{rgb}{0.88,1,1}
\definecolor{LawnGreen}{rgb}{0.48,0.98,0}
\definecolor{mygreen}{RGB}{28,172,0} 
\definecolor{mylilas}{RGB}{170,55,241}
\definecolor{lightPaleGreen}{RGB}{152,251,152}
\definecolor{fullPaleGreen}{RGB}{173,255,47}
\definecolor{lightPaleRed}{RGB}{240,128,128}
\definecolor{fullPaleRed}{RGB}{255,99,71}

\usepackage{array}
\usepackage{ragged2e}
\newcolumntype{P}[1]{>{\RaggedRight\hspace{0pt}}p{#1}}
\newcolumntype{R}[1]{>{\RaggedLeft\arraybackslash}p{#1}}

\IEEEoverridecommandlockouts

\begin{document}

\title{The Robotarium: A remotely accessible swarm robotics research testbed}

\author{Daniel~Pickem,
	Paul~Glotfelter,
	Li~Wang,
	Mark~Mote,
	Aaron~Ames,
	Eric~Feron,
	and~Magnus~Egerstedt
\thanks{*This research was sponsored by grants No. 1531195 and 1544332 from the U.S. National Science Foundation.} 
\thanks{The authors are with the Georgia Institute of Technology, Atlanta, GA 30332, USA, \{daniel.pickem,paul.glotfelter,liwang,mmote3,ames,feron,\newline magnus\}@gatech.edu.}%
}

\maketitle

\begin{abstract}
This paper describes the \textit{Robotarium} -- a remotely accessible, multi-robot research facility. The impetus behind the Robotarium is that multi-robot testbeds constitute an integral and essential part of the multi-robot research cycle, yet they are expensive, complex, and time-consuming to develop, operate, and maintain. These resource constraints, in turn, limit access for large groups of researchers and students, which is what the Robotarium is remedying by providing users with remote access to a state-of-the-art multi-robot test facility. 
This paper details the design and operation of the Robotarium and discusses the considerations one must take when making complex hardware remotely accessible. In particular, safety must be built into the system already at the design phase without overly constraining what coordinated control programs users can upload and execute, which calls for minimally invasive safety routines with provable performance guarantees.
\end{abstract}

\section{Introduction}
\label{sec:introduction}
Coordinated control of multi-robot systems has received significant attention during the last decades, with a number of distributed control and decision algorithms being developed to solve a wide variety of tasks, ranging from environmental monitoring to collective material handling. These developments have been driven by a confluence of algorithmic advances, increased hardware miniaturization, and cost reduction, and a number of compelling multi-robot testbeds have been developed  (e.g., \cite{Johnson2006, Casan2015}). However, despite these advances, it is still a complex and time-consuming proposition to go from theory and simulation, via a small team of robots, all the way to a robustly deployed, large-scale multi-robot system.
Yet, actual deployment is crucial to advance multi-robot research since it is increasingly difficult to faithfully simulate all the issues associated with making multiple robots perform coordinated tasks. Even more difficult is the discovery of new issues based on analytical models of multi-robot systems alone.

\begin{figure}[tbp]
  \begin{center}
    \includegraphics[width=0.45\textwidth]{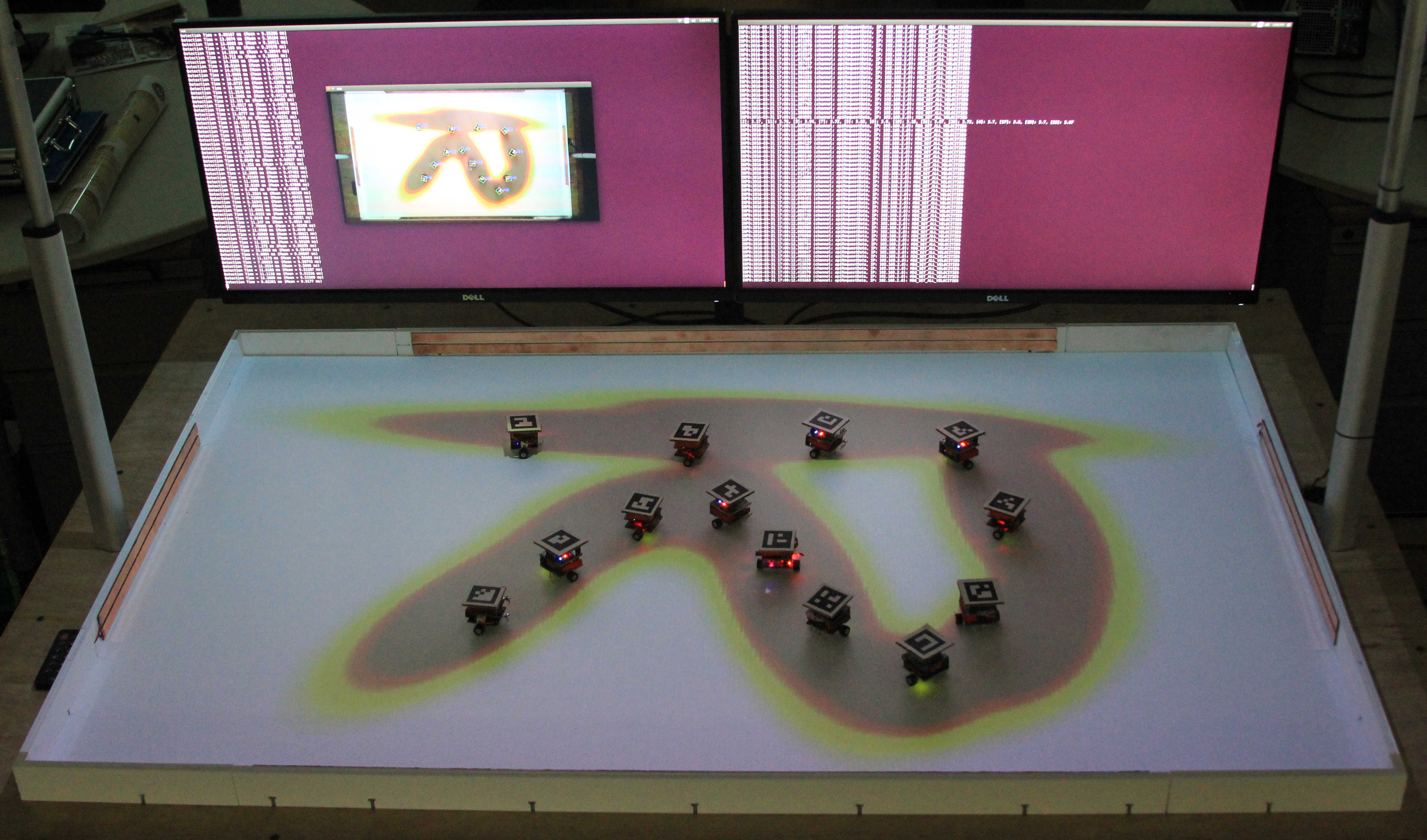}
    \caption{Example of a coverage control algorithm executed on the Robotarium using 13 GRITSBot robots. The desired density function is projected onto the testbed arena in the shape of the letter R.}
    \label{fig:static_coverage_control_title_page}
  \end{center}
\end{figure}

The Robotarium, as shown in Figure \ref{fig:static_coverage_control_title_page}, is an open, remote-access multi-robot testbed, explicitly designed to address this theory-simulation-practice gap by providing access to a testbed that is flexible enough to allow for a number of different scientific questions to be asked, and different coordination algorithms to be tested. 
What sets the Robotarium apart from other testbeds is its explicit focus on supporting \textit{safe remote-access multi-robot research}, as opposed to testbeds that can be accessed remotely but do not explicitly consider formal safety measures or have an educational, as opposed to a research focus. Throughout this work, we will interpret safety as the avoidance of damaging collisions and quantify it through safety scores that determine whether user code is allowed to execute without further safety measures or not. The Robotarium makes hardware remotely accessible to both trusted and untrustworthy or malicious users while avoiding damage in a provable way. 
In this paper, we discuss how the Robotarium is structured and, in particular, how the explicit focus on being a \textit{flexible} and \textit{safe} remote-access research platform informs the design.\footnote{The report from a recent NSF Workshop on Remotely Accessible Testbeds \cite{Egerstedt2015} identified this inherent safety/flexibility tension as one of the key questions when pursuing a "science of remote access".}

\section{Related Work}
\label{sec:relatedWork}
In this section, we briefly survey the field of remote access testbeds that have been successful in their respective domains and broadly categorize them along the following dimensions: multi-robot testbeds, sensor network testbeds, and remotely accessible educational tools. A comprehensive overview can be found in \cite{Groeber2007,Jimenez-Gonzalez2013,Tonneau2015} and the references therein.
  
\subsection{Multi-Robot Testbeds}
Numerous remotely accessible multi-robot testbeds with a focus on robot mobility have been proposed over the years - for example the Mobile Emulab \cite{Johnson2006}, or the HoTDeC testbed \cite{Vladimerou2006}. A comprehensive list of multi-robot testbeds can be found in \cite{Jimenez-Gonzalez2013}.
Generally speaking, testbeds in this domain contain robots that occupy a significantly larger footprint and are more expensive than the Robotarium robots, which is an inherent obstruction to using large numbers of robots.\footnote{A GRITSBot can be built for approximately \$60 or bought pre-assembled for approximately \$100 - see the bill of materials at \url{www.robotarium.org}} 
%
The main difference between these testbeds and the Robotarium, however, is that the Robotarium explicitly addresses the robot safety aspect of remote accessibility such that provable damage avoidance of the Robotarium's physical assets is guaranteed even with untrustworthy or malicious users. The testbeds mentioned above in principle allow remote access for cases where a user can be trusted to not damage the hardware but do not explicitly address the safety issues involved once users have been approved for use. Unlike these testbeds, the Robotarium is inherently safe to operate because built-in (online and offline) safety measures prevent users from causing accidental or purposeful damage to the robots. 

\subsection{Sensor Networks Testbeds and Cybersecurity}
Some of the earliest remote-access testbeds reside in the sensor networks and cybersecurity domains. Limiting access to largely immobile computing and sensing nodes mitigated the risk of making them publicly accessible. This category includes testbeds such as ORBIT \cite{Raychaudhuri2005}, FIT IoT-Lab \cite{Adjih2015}, DeterLab \cite{Mirkovic2012} and CONET \cite{Martinez-deDios2013}, with the ORBIT testbed \cite{Raychaudhuri2005} and the DeterLab \cite{Mirkovic2012} standing out for their over a decade long remote operation.
Of these testbeds, the FIT IoT-Lab \cite{Adjih2015} and CONET \cite{Martinez-deDios2013} contain mobile sensing nodes which both require significant resources and spaces to operate. The FIT IoT-Lab's over 200 mobile nodes are spread across six sites in France. Compared to the Robotarium's focus on mobility and coordinated control, FIT Iot-Lab's main focus, however, lies on communications and networking research in an IoT context.
For a comprehensive overview of networking testbeds refer to \cite{Tonneau2015} and the references therein.

\subsection{Educational Testbeds}
A number of testbeds have originated in the educational domain. For example, the Robotic Programming Network (RPN) \cite{Casan2015} makes a single humanoid robot remotely accessible while Robotnacka \cite{Petrovic2012} provides access to three mobile robots. A comprehensive overview of other educational testbeds can be found in \cite{Groeber2007}. While providing the required infrastructure for remote access, compared to the Robotarium, most educational testbeds contain small numbers of robots and are not explicitly designed to be research platforms for multi-robot or swarm robotics experiments.

\section{The Robotarium}
\label{sec:robotarium}
The Robotarium  is a swarm-robotic research testbed that is accessible through a public web interface and gives users the flexibility to test a variety of multi-robot algorithms (see examples of remote experiments in Section \ref{subsec:external_users}). In particular the Robotarium tackles the challenge of robust, long-term, and safe operation of large groups of robots with minimal operator intervention and maintenance. The continuous operation of the Robotarium highlights the need for automated maintenance, which relies on robust position tracking, automated battery recharging, and provably collision-free execution of motion paths. In this section, we will outline how these requirements guided the design of the first Robotarium instantiation, and elaborate on both the hardware and software architectures of the Robotarium.

\begin{figure*}[tbp]
  \begin{center}
    \includegraphics[width=0.85\textwidth]{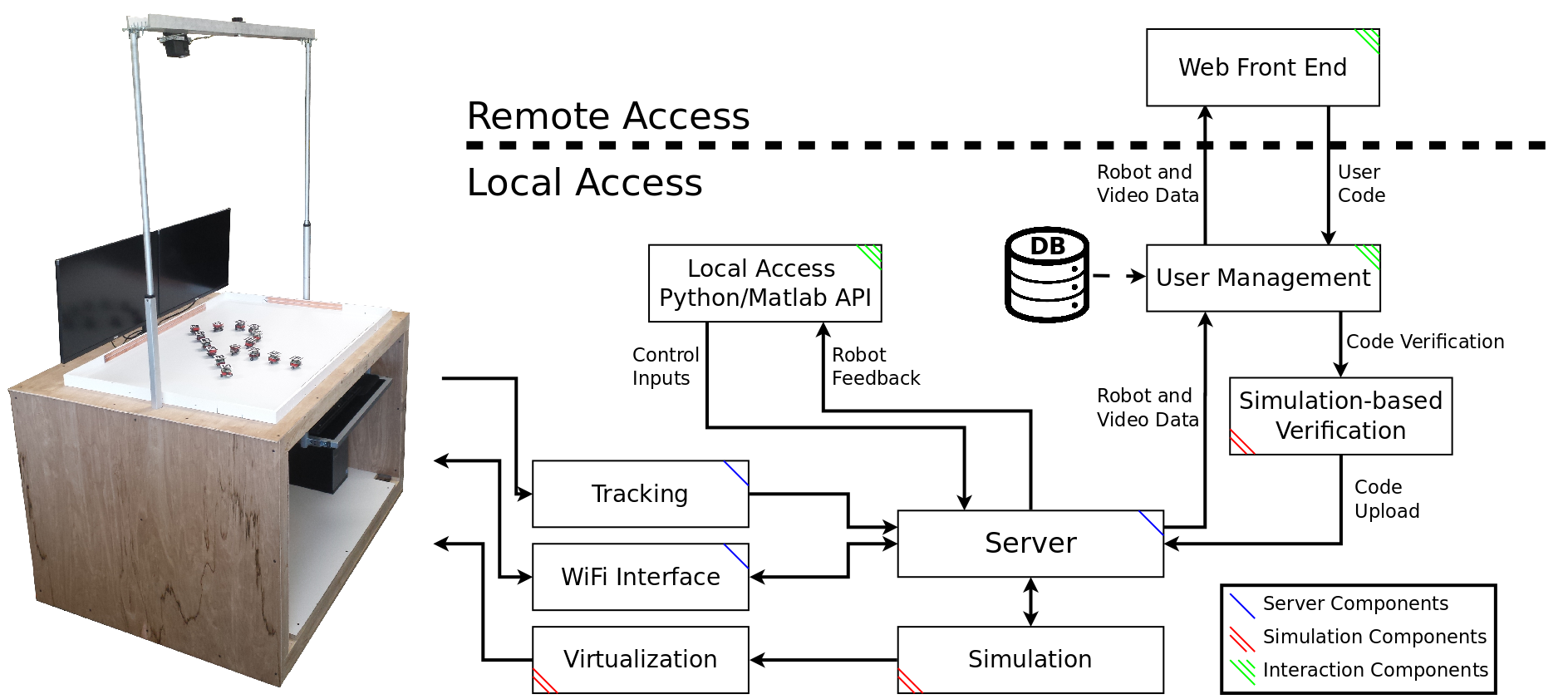}
    \caption{System architecture overview. The Robotarium includes components that are executed locally on Robotarium infrastructure as well as user-facing components that run on remote user machines (web front end). Three components interact directly with the robot hardware -- tracking, wireless communication, and virtualization. The remaining components handle user management, code verification and upload to the server, as well as coordination of user data and testbed-generated data.}
    \label{fig:system_architecture}
  \end{center}
\end{figure*}

\subsection{Design Considerations}
\label{subsec:design_considerations}
As a shared, remotely accessible, multi-robot facility, the Robotarium's main purpose is to lower the barrier of entrance into multi-agent robotics. Therefore, for the Robotarium to fulfill its intended use effectively, it has to implement a number of high-level design requirements aimed at accessibility, ease of maintenance, intuitive interaction, and safe and secure code execution.

\begin{itemize}
    \item Enable inexpensive replication of the Robotarium through low-cost, open-source robots (currently up to 20 robots are available).
    \item Enable intuitive interaction with and simple data collection from the Robotarium through a public web interface that facilitates code submission and data/video retrieval.
    \item Enable a seamless switch between development in simulation and execution on the physical robots facilitated by a data-driven characterization of the simulation-hardware gap of the robots.
    \item Minimize the cost and complexity of maintaining a large collective of robots through convenience features such as automatic charging and tracking.
    \item Integrate safety and security measures to protect the Robotarium from damage and misuse through guaranteed collision avoidance.
\end{itemize}

\subsection{Prototype Design}
\label{subsec:prototype_design}
This section elaborates on the hardware and software components of the Robotarium as well as their interaction with each other, the robots, and the users (see Fig. \ref{fig:system_architecture}). The Robotarium hardware includes the robots themselves, the position tracking system, the wireless communication hardware, as well as a charging system built into surface of the arena. The software back end consists of the simulation and virtualization infrastructure (also used for simulation-based code verification), interaction components (APIs), and the coordinating server application. 

\subsubsection{Hardware}
\label{subsubsec:hardware}
The Robotarium is meant to provide a well integrated, immersive user experience with a small footprint (the current testbed measures $130 \times 90 \times 180$ cm), and features that allow a large swarm to be maintained effortlessly. At the core of the Robotarium are our custom-designed robots - the GRITSBots (introduced in \cite{Pickem2015}). These inexpensive, miniature differential drive robots simplify operation and maintenance of the Robotarium through features such as (i) automated registration with the server and overhead tracking system, (ii) automatic battery charging, and (iii) wireless (re)programming.
While the initial implementation of these features has been described in detail in \cite{Pickem2015}, major design revisions warrant a review and update.\footnote{A detailed description of the robot's hardware design including its design and code files are open-source and can be downloaded at \url{https://github.com/robotarium/GRITSBot_hardware_design}.}
    
\begin{figure}[tbp]
	\centering
	\subfloat[3D rendering of the GRITSBot]{\includegraphics[width=0.23\textwidth]{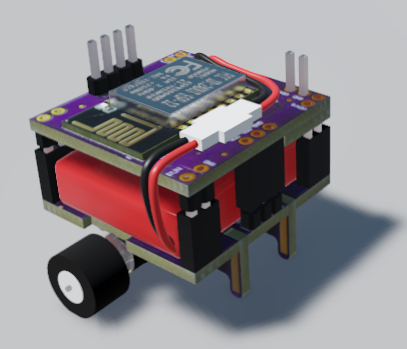} \label{fig:robot_layers}}
	\hfil{}
    \subfloat[A GRITSBot wirelessly charging on a base station of the Robotarium.]{\includegraphics[width=0.23\textwidth]{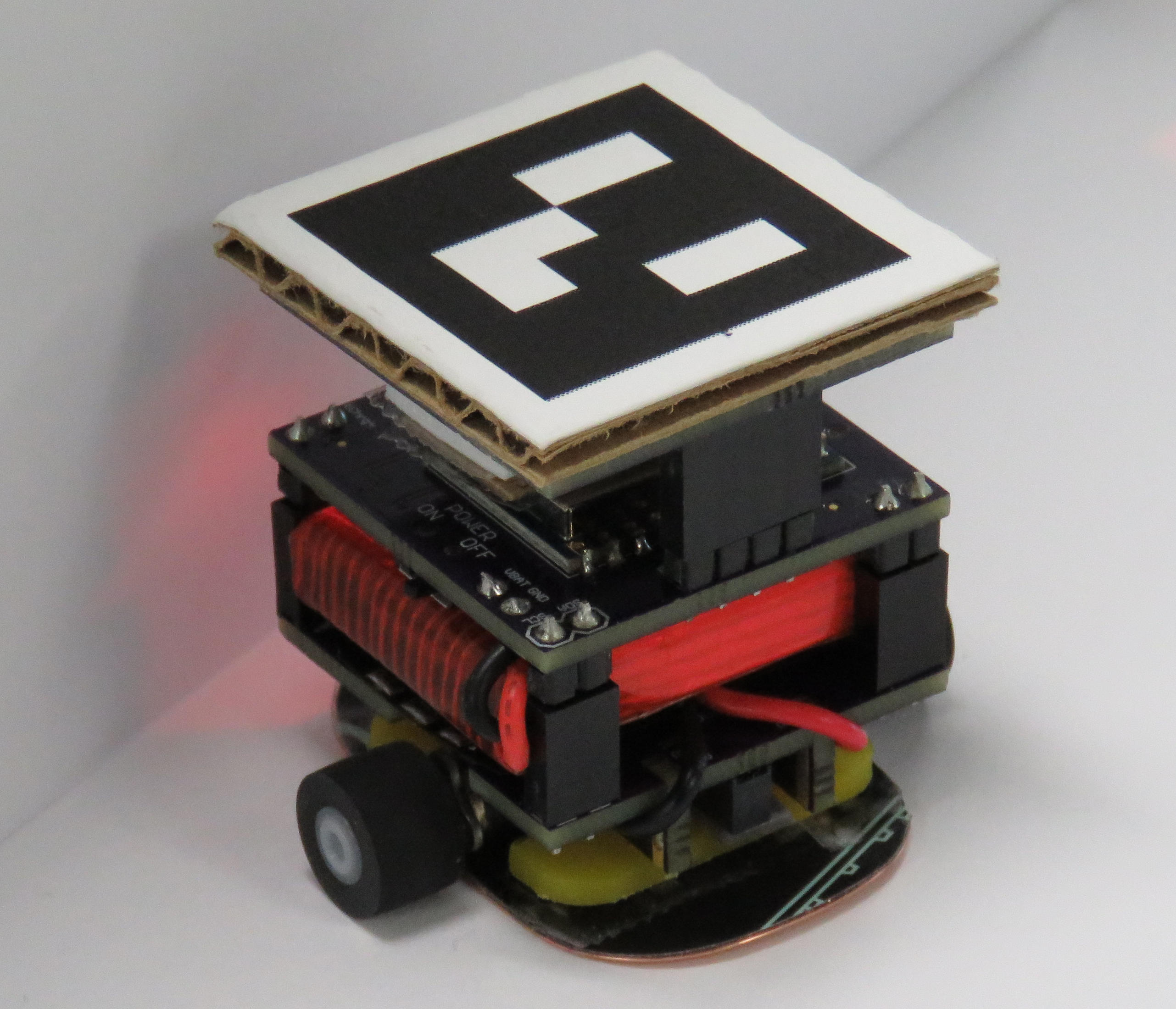} \label{fig:robot_charging}}
	\caption{3D model and current hardware implementation of the GRITSBot.}
    \label{fig:robots}
\end{figure}

\begin{itemize}
\item{\textit{Robots:}}
The GRITSBot's modular design consists of a main board handling high-level intelligence, connectivity, power conditioning, and charging as well as a motor board responsible for the robot's motion. Compared to the initial specifications in \cite{Pickem2015}, the GRITSBot has undergone multiple design iterations. Specifically, the robot's main board has been upgraded with the WiFi-enabled ESP8266 chip popular in the Internet of Things community. Operating at 160 MHz, this chip is capable of handling wireless communication, pose estimation, low-level control of the robot, as well as high-level behaviors. The ESP8266 chip's WiFi transceiver supports the IEEE 802.11 B/G/N standards with a bandwidth of up to 54 MBit/s. Note that the ESP8266 chip also enables wireless over-the-air code upload to the robots, which allows the Robotarium back end to upload new firmware to the robots within seconds. Equipped with a 400 mAh LiPo battery, the robot is capable of operating up to 40 minutes on a single charge while recharging takes approximately 45 minutes.
The GRITSBot's motor board features an Atmega 168 microcontroller, which controls the stepper motors. Compared to the design shown in \cite{Pickem2015}, the motor board has been equipped with additional introspective sensors (specifically motor current and temperature sensors) to enable predictive diagnostics of the robot's hardware state. Note that the robots are currently not equipped with sensor boards since distance sensing can be emulated through the back end server which tracks the positions of all robots.
    
\item{\textit{Tracking:}}
The global position of all robots is retrieved through an overhead tracking system and is required to close the position control feedback loop. The Robotarium uses a single webcam in conjunction with ArUco tags for tracking.\footnote{ArUco is an OpenCV-based library for Augmented Reality applications.} Note that most decentralized algorithms do not rely on global position updates but rather sensor data. However, system maintenance such as recharging robots automatically or setting up an experiment relies on global position data. 
      
\item{\textit{Charging:}}
A crucial component of a self-sustaining testbed is an automatic recharging mechanism for its robots. The GRITSBot has been equipped with a wireless charging system for autonomous charging through a receiver coil attached to the robot (see Fig. \ref{fig:robot_charging}) and transmitters built into the Robotarium arena surface (both devices rely on the Qi wireless charging standard). 
Automatic recharging of robots is an essential aspect that will enable the long-term use of robots and the automated management of the Robotarium hardware with minimal operator intervention (see Section \ref{subsec:long_term_operation}) and at the same time make the continuous operation of the Robotarium economically feasible. 
\end{itemize}

\subsubsection{Software}
\label{subsubsec:software}
The software components managing the operation of the Robotarium can be broadly grouped into three categories: simulation-based components, components enabling the interaction with the testbed, and coordinating server applications (see Fig. \ref{fig:system_architecture}).
  
\begin{itemize}
\item{\textit{Simulation:}}
The simulation capabilities of the Robotarium are leveraged in three distinct ways: prototyping of user code, verification of user-provided code, and adding virtual robots. The simulators enable users to prototype and test their algorithms on their own machines before submitting them for execution on the Robotarium.\footnote{The simulators are currently implemented in Matlab and Python and available in the 'Downloads' section at \url{http://robotarium.org/}} Once submitted, but before being executed on the testbed, the same simulation infrastructure verifies collision-free execution of user code (see Section \ref{subsec:sim_based_verification}). Additionally, the simulator also allows adding virtual robots to the testbed arena that are capable of interacting with the physical robots through the server back end.
    
\item{\textit{Interaction:}}
These components govern how users can interact with the Robotarium. Two principal methods of interaction are enabled by the components shown in Fig. \ref{fig:system_architecture}: local access through provided APIs as well as remote interaction through web-based code upload.\footnote{Note that no real-time teleoperation of robots is enabled for security reasons. Submitted user-code is executed locally on the Robotarium server and as such latency is negligible.} Local access requires users to connect to the Robotarium's WiFi network and is primarily used for development purposes. Remote access, on the other hand, requires users to implement and test algorithms in simulation before submitting them to the Robotarium via its web interface at \url{www.robotarium.org}. Submitted code undergoes simulation-based verification that checks for error- and collision-free execution before code is executed on physical robots (see Section \ref{subsec:sim_based_verification}).\footnote{Remote access to the Robotarium requires manual screening of applicants. Interested users can apply for access privileges via the website \url{www.robotarium.org}.} 
  
\item{\textit{Coordination:}}
The server application is the central coordinating instance responsible for executing user code, routing commands and data to and from robots, transmitting global position data to the robots, and managing simulated virtual robots. In addition the server logs all generated data and makes them available to users. 
Note that the robots execute a velocity controller onboard. However, the user's control code is executed on the server and essentially remote controls the robots by providing velocity control inputs to the robots. This centralized execution increases overall robustness of the Robotarium, simplifies data logging as well as establishing formal safety guarantees, and facilitates automatic maintenance.\footnote{Note, however, that other decentralized communication architectures such as peer-to-peer communication are also enabled by the WiFi-based communication architecture.} Note that centralization can lead to communication and computation bottlenecks. Given the total bandwidth of 802.11g WiFi of 54 MBit/s and typical bandwidth requirements of 3 KBytes/s/robot the theoretical upper limit is 18,000 robots. Clearly, WiFi collisions, buffering issues, interference, and other issues will practically lower that number. However, operating hundreds of robots simultaneously is well within the limits of the Robotarium's WiFi-based communication architecture. Computational scalability is addressed in Section \ref{subsec:safety_barrier_certificates}.
\end{itemize}

\section{Safety}
\label{sec:safety}
Allowing remote users to control the Robotarium's physical equipment poses inherent risks to the integrity and safety of the hardware. To avoid damaging the hardware, a combination of offline simulation-based verification and online collision avoidance using barrier certificates is employed.
By default, the execution of all user-supplied control code will be safe-guarded using barrier functions (see Section \ref{subsec:safety_barrier_certificates}). However, users can bypass this online safety mechanism if they achieve a sufficiently high safety score during the offline simulation-based verification step (see Section \ref{subsec:sim_based_verification}).

\subsection{Simulation-based Verification}
\label{subsec:sim_based_verification}
In this section we introduce an offline method to characterize the \textit{safety} of a particular experiment and its suitability for deployment on the Robotarium. A \textit{safety score} measures the frequency and severity of collisions in simulation to predict the collision behavior of an experiment once executed on hardware. This method relies on stochastic Monte Carlo simulations to compute the expected values of damage and safety over multiple simulation runs (currently 50). Uncertainties in the robots' dynamics model, their initial positions, as well as the observation model of the overhead camera are accounted for through added Gaussian noise. 
Note that strict collision-free execution is not enforced but minor collisions below certain damage thresholds are allowed. A sufficiently high safety score of an experiment allows execution on the Robotarium without additional online safety mechanisms (see Section \ref{subsec:safety_barrier_certificates}).

To formalize safety scores, let the index set of $N$ robots be $\mathcal{M}=\{1,2,...,N\}$. Then the function $D\in\mathbb{R}^{+}$ computes the cumulative $damage$ done to $N$ robots over a time horizon $T$ and is defined as follows.
\[
D=\sum_{i\in\mathcal{M}}D_{i}\triangleq\sum_{i\in\mathcal{M}}\int_{0}^{T} I_i(t) \delta_{i}(t)dt,
\]
Here, $\delta_i(t)$ captures the rate at which robot $i$'s kinetic energy is lost through a collision at time $t$ and $I_i(t)$ is an indicator function that is $0$ if robot $i$ is not colliding with any other object at time $t$ and $1$ otherwise. In discrete time $D_i$ can be approximated using the work-energy principle as follows.
\begin{equation}
D_i \approx \sum_{k = 0}^{T/\Delta t - 1} I_i(k \Delta t) \frac{m}{2} \left[ v^2(k \Delta t) - v^2((k+1) \Delta t)\right],
\label{eqn:damage_approximation}
\end{equation}
where $m=60g$ is the mass of the robot, $\vec{v}$ is its velocity vector such that $v^2 = \vec{v} \cdot \vec{v}$, and $\Delta t = 1/30s$ is the time step of the system governed by the tracking camera update rate. Note that Eqn. \eqref{eqn:damage_approximation} assumes that a reduction in robot $i$'s velocity is proportional to a loss in kinetic energy and therefore approximates damage.\footnote{The maximum energy loss possible from an inelastic collision involving a single GRITSBot weighing at most 60g directed towards a wall at its maximum speed of 0.1 m/s is 9.9 $\mu J$.}
The cumulative \textit{safety score} $S$ is then defined as, 
\[
S\triangleq1-\frac{D}{D_{max}},
\]
where $D_{max}$ is the maximum allowable damage threshold for the entire experiment. Note that $S$ is a unit-less value $S\in(-\infty,1]$ which captures the energy lost from collisions over the time span of the experiment. In a similar fashion, we can limit the allowable damage done to any one robot $i$ in the swarm by defining an individual safety score $s_i\triangleq1-\frac{D_i}{d_{i,\mathrm{max}}} \in(-\infty , 1]$  with respect to a separate threshold $d_{i,\mathrm{max}} \in\mathbb{R}^+$. Experiments are allowed to proceed for unmodified deployment in the Robotarium when both the cumulative and individual safety scores are positive. 

\begin{figure*}[t]
	\centering
	\subfloat[Time 2.67s]{\includegraphics[width=0.3325\textwidth]{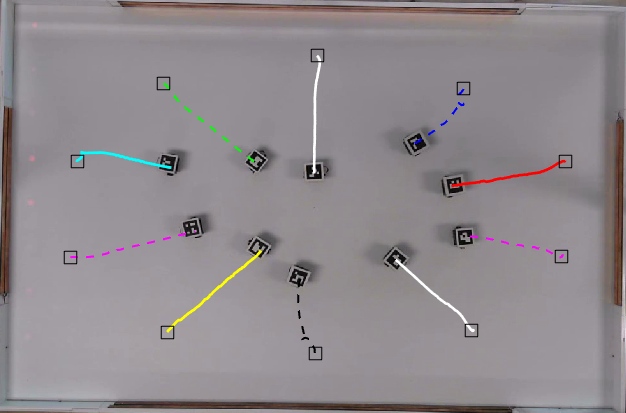} \label{fig:exp2}}
	\subfloat[Time 9.50s]{\includegraphics[width=0.325\textwidth]{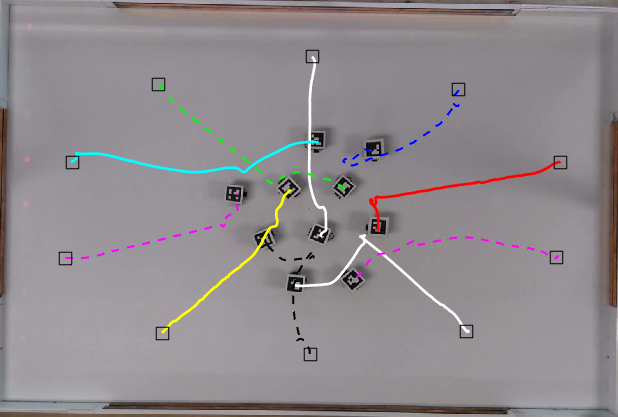} \label{fig:exp3}} 
	\subfloat[Time 19.83s]{\includegraphics[width=0.325\textwidth]{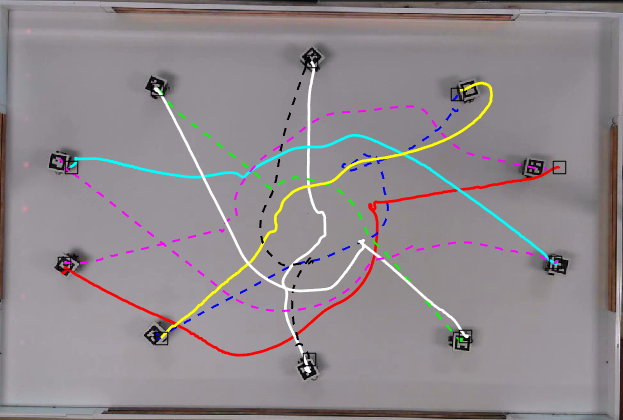} \label{fig:exp5}}
	\caption{Ten GRITSBots swap positions with active safety barrier certificates. The robots' trajectories are shown together with square markers representing their initial positions. \label{fig:expbarrier}}
\end{figure*}

\subsection{Safety Barrier Certificates}
\label{subsec:safety_barrier_certificates}
The Robotarium uses \textit{Safety Barrier Certificates} to guarantee provably collision-free behavior of all robots, which ensures the following three principles.
\begin{itemize}
  \item All robots are provably safe in the sense that collisions are avoided.
  \item Users' commands are only modified when collisions are imminent.
  \item Collision avoidance is executed in real-time (in excess of 30 Hz update rate).
\end{itemize} 
Safety barrier certificates are enforced through the use of control barrier functions, which are Lyapunov-like functions that can provably guarantee forward set invariance, i.e. if the system starts in the safe set, it stays in the safe set for all time. A specific class of maximally permissive control barrier functions was introduced in \cite{ames2014}, whose construction provides the basis for the minimally invasive safety guarantees afforded by the Robotarium.

Consider a team of $N$ mobile robots with the index set $\mathcal{M}=\{1,2,...,N\}$. Each robot $i$ uses single integrator dynamics of the form $\dot{x}_i = u_i$, where $x_i\in\mathbb{R}^2$ is the planar position of robot $i$, and $u_i\in\mathbb{R}^2$ is its input velocity.\footnote{Single integrator dynamics can be easily mapped to the GRITSBot's unicycle dynamics using a nonlinear inversion method. It is also important to note that safety barrier certificates can be extended to more complex dynamical systems as well.}
Additionally, robot $i$'s velocity $u_i$ is bounded by $\|u_i\|\leq \alpha, \forall i\in\mathcal{M}$. Let $x=[x_1^T, x_2^T, ..., x_N^T]^T$ and $u=[u_1^T, u_2^T, ..., u_N^T]^T$ denote the aggregate state and velocity input of the entire team of robots. 
To avoid inter-robot collisions, any two robots $i$ and $j$ need to maintain a minimum safety distance $D_s$ between each other. This requirement is encoded into a pairwise safe set $\mathcal{C}_{ij}, ~\forall~ i\neq j$, which is a super level set of a smooth function $h_{ij}(x)$, 
\begin{equation}\label{eqn:setcij}
\mathcal{C}_{ij} = \{x_i\in\mathbb{R}^2 ~|~h_{ij}(x)=\|x_i-x_j\|^2-D_s^2\geq 0\}.
\end{equation}

The function $h_{ij}(x)$ is called a control barrier function, if the admissible control space
\begin{equation}\label{eqn:setkij}
K_{ij}(x) = \left\{u\in\mathbb{R}^{2N} \;\middle|\; \frac{\partial h_{ij}(x)}{\partial x}u\geq -\gamma h_{ij}(x)\right\},
\end{equation}
is non-empty for all $x_i\in\mathcal{C}_{ij}$. It was shown in \cite{xu2015} that if the control input $u$ stays in $K_{ij}(x)$ for all time, then the safe set $\mathcal{C}_{ij}$ is forward invariant. In addition, the forward invariance property of $\mathcal{C}_{ij}$ is robust with respect to different perturbations on the system. 

Combining \eqref{eqn:setcij} and \eqref{eqn:setkij} as well as the single integrator dynamics, the velocity input $u$ needs to satisfy
\begin{equation*}
-2(x_i-x_j)u_i+2(x_i-x_j)u_j\leq \gamma h_{ij}(x),~ \forall~ i\neq j.
\end{equation*}
This inequality can be treated as a linear constraint on $u$ when the state $x$ is given, i.e., 
$
A_{ij}u\leq b_{ij}, ~\forall~ i\neq j,
$
where 
\begin{eqnarray}
  A_{ij} &=&[0, \ldots, \underbrace{-2(x_i-x_j)^T}_{\text{robot} ~i}, \ldots, \underbrace{2(x_i-x_j)^T}_{\text{robot} ~j}, \ldots, 0 ] \nonumber\\ 
  b_{ij} &=& \gamma h_{ij}(x) \nonumber.
\end{eqnarray}
Similar constraints must be established for the workspace boundary. The corresponding safety set of robot $i$ with regards to the boundary is denoted by $\bar{\mathcal{C}}_i$, and the corresponding constraints by $\bar{A}_{i}u_i\leq \bar{b}_{i}, ~\forall~i\in\mathcal{M}$.

Combining these constraints --  \emph{all} pairwise collisions and collisions with the workspace boundaries -- results in the safety set for the entire team as
\begin{equation*}
\mathcal{C} = \underset{{i \in \mathcal{M}}}{\prod} \Big\{{\bigcap_{\substack{j \in \mathcal{M}\\ j\neq i}}\mathcal{C}_{ij} \bigcap\bar{\mathcal{C}}_{i} }\Big\}. 
\end{equation*}
The forward invariance of the safe set $\mathcal{C}$ is guaranteed by the safety barrier certificates, which are defined as
\begin{equation}\label{eqn:certificates}
K(x) = \left\{u\in\mathbb{R}^{2N}\;\middle|\;A_{ij}u\leq b_{ij}, \bar{A}_{i}u_i\leq \bar{b}_{i}, ~\forall~ i\neq j \right\}.
\end{equation}
These safety barrier certificates define a convex polytope $K(x)$ in which safe control commands must stay. By constraining users' control commands to within $K(x)$, the Robotarium is guaranteed to operate in a provably collision-free manner.

The \textit{minimally invasive} nature of  barrier certificate-enabled collision avoidance stems from the fact that the deviation between the user-specified control signal and the actual, safe, executed signal is minimized, subject to the safety constraints through a Quadratic Program (QP)-based controller
\begin{equation} \label{eqn:QPcontrol}
 \begin{aligned}
u^* =  & \:\: \underset{u\in\mathbb{R}^{2n}}{\text{argmin}}
 & & J(u) = \sum_{i=1}^{N} \|{u}_{i} - \hat{u}_{i} \|^2  &\\
 & \qquad \text{s.t.}
 & & A_{ij}u \leq b_{ij}, \qquad  &\forall~ i\neq j,  &  \\
 &
 & & \bar{A}_{i}u_i\leq \bar{b}_{i},\: \qquad  &\forall~ i\in \mathcal{M}, & \\
 &
 & &    \| u_i\|_\infty  \leq \alpha, \qquad &\forall~ i \in\mathcal{M},  
 \end{aligned}
\end{equation}
where $\hat{u}$ is the user's control command, $u^*$ is the actual control command, and $\alpha$ is the bound for the control input. Note that in the absence of impending collisions (i.e. when the safety barrier certificates in \eqref{eqn:certificates} are satisfied), the user's code is executed faithfully. When violations occur, a closest possible (in a least-squares sense) safe control command is computed and executed instead. An experiment showing ten GRITSBots swapping positions with active safety barrier certificates is shown in Fig. \ref{fig:expbarrier}, while the corresponding video is referenced in Table \ref{table:multimedia_extensions}.

\subsubsection*{Scalability of Safety Barrier Certificates} Safety barrier certificates are computed in a centralized fashion on the Robotarium's back end server and therefore scalability is a concern.
As the size of the swarm increases, the number of decision variables ($u$) in the QP-based controller increases linearly, while the number of pairwise safety constraints grows quadratically. However, a more computationally efficient implementation similar to \cite{Borrmann2015} is possible, where agent $i$ only considers its neighbors for collision avoidance and the certificates computation can be distributed to individual agents. 
More specifically, the robot's finite physical dimensions limit the maximum robot density. For example, for a minimum safety distance of $D_s=8cm$ and a neighborhood radius of $20cm$, any given neighborhood can contain at most 26 other robots, which limits the size of each individual robots QP problem to 2 decision variables and at most 26 linear constraints.
The computation time of safety barrier certificates for centralized as well as decentralized computation is shown in Table \ref{table:qptime}. Note that the decentralized barrier certificates were computed on a single central computer.\footnote{Barrier certificates were computed on an Intel I7 4790 3.6 GHz with 16 GB of memory.} Thus, the total computation time $T_d$ is divided by $N$ to characterize the decentralized and fully parallel implementation. As Table \ref{table:qptime} shows, while centralized computation suffers from scaling up the number of robots, the computation time of decentralized safety barrier certificates remains below 10ms even for 100 robots. In fact, decentralized safety barrier certificates can handle 100 GRITSBots with an update frequency of 185Hz and therefore scale to large numbers of robots without compromising update rates.

\section{Usage}
\label{sec:usage}
This section highlights the main usage features of the Robotarium: continuous operation and auto-recharging of robots, safe remote access for external users, and the characterization and closing of the simulation-hardware gap that can prevent the successful execution of controls code on physical robots. The following sections detail each of these features.

\subsection{Long-term Operation}
\label{subsec:long_term_operation}
The robust long-term operation of the Robotarium is predicated on a reliable and autonomous charging mechanism for its robots. In this section we therefore provide experimental evidence of the reliability of the GRITSBot's wireless charging mechanism (see Section \ref{subsubsec:hardware}) and the Robotarium's capabilities for continuous operation. 

This section summarizes the results of three experiments, each of which made use of three robots and continuously executed an example algorithm over the course of 140, 148, and 240 minutes, respectively. During these times, each robot executed 37, 39, and 60 autonomous recharge cycles, i.e. successfully charged its battery through the wireless chargers.\footnote{Note that we shortened each charge cycle in the interest of time and therefore on average a charge cycle only lasted approximately 6 minutes - 3 minutes of experiment and 3 minutes of charging. As mentioned in Section \ref{subsubsec:hardware} the GRITSBot's runtime on a single charge is up to 40 minutes.} Out of a total of 111, 117, and 180 autonomous charge cycles, manual intervention was only required twice in the first experiment and once in the second, because a robots onboard control software froze. The success rate of the autonomous charging system was therefore $98.1$\%, $99.1$\%, and $100$\%, respectively.
A time-lapse video showing the continuous operation of three robots is referenced in Table \ref{table:multimedia_extensions}.

\subsection{External Users}
\label{subsec:external_users}
This section presents a selection of external user experiments that have been developed using a Robotarium-provided simulator and executed on the Robotarium using the software infrastructure shown in Section \ref{subsubsec:software} and Fig. \ref{fig:system_architecture}. These examples were chosen as representative samples since they highlight the breadth of algorithms that can be executed on the Robotarium but also validate its remote access aspect. Note that the corresponding videos are referenced in Table \ref{table:multimedia_extensions}.

\begin{figure*}[t]
	\centering
	\subfloat[Distributed formation control of six GRITSBots assembling a regular hexagon starting at random initial positions.]{\includegraphics[width=0.3\textwidth]{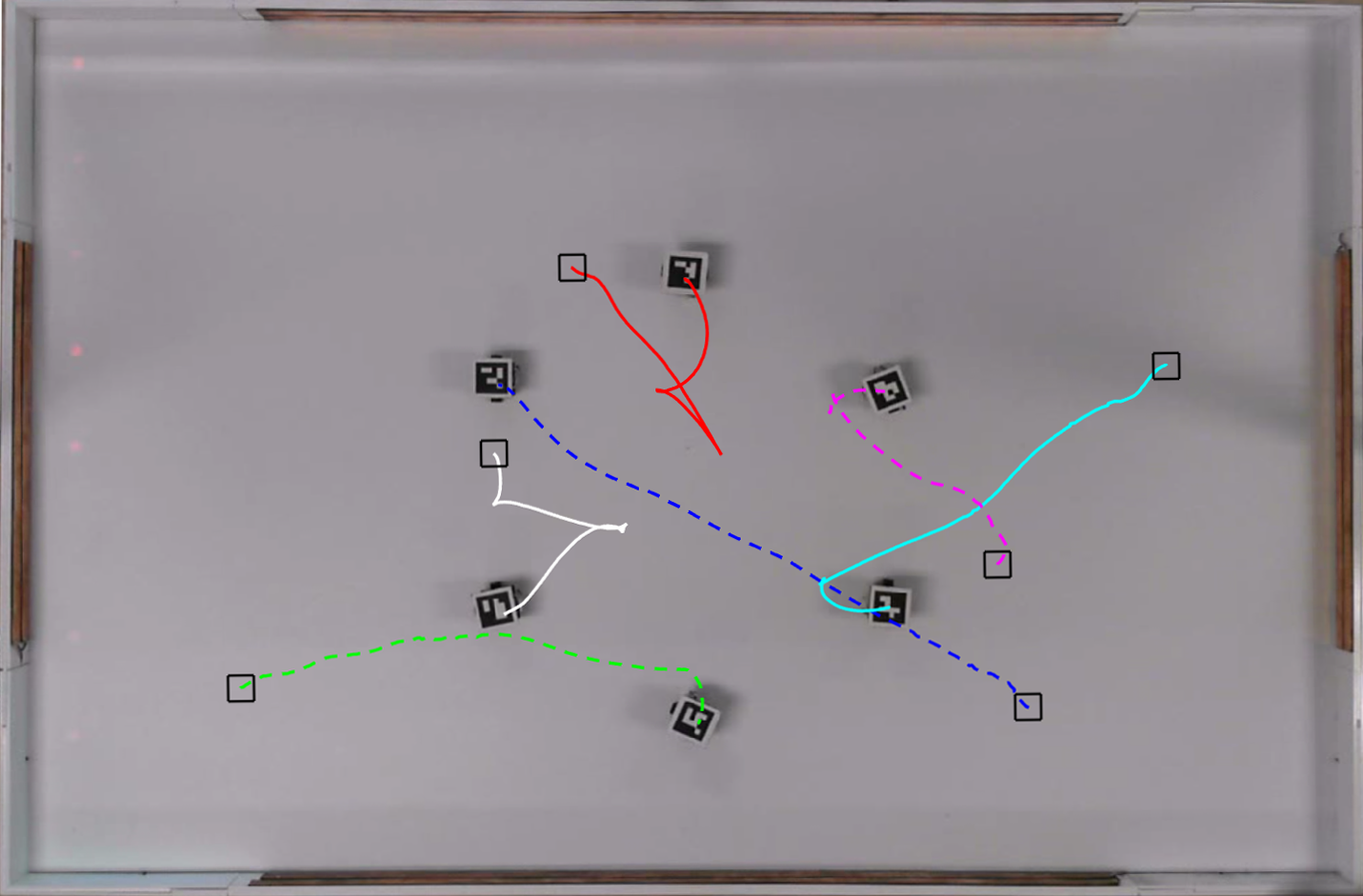} \label{fig:external_users_formation_control}}
    \qquad
	\subfloat[Fault-tolerant consensus with five collaborating robots (shown in blue) and one malicious robot (shown in red). The collaborative agents achieve rendezvous despite the malicious robot's efforts to prevent it.]
	{\includegraphics[width=0.29\textwidth]{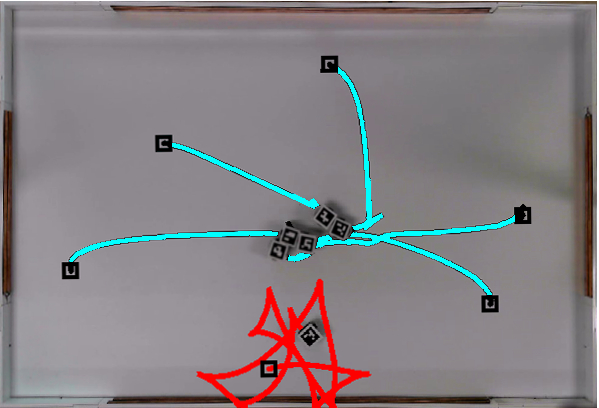} \label{fig:external_users_consensus}} 
    \qquad
	\subfloat[Passivity-based attitude synchronization implemented on a team of nine GRITSBots. ]{\includegraphics[width=0.295\textwidth]{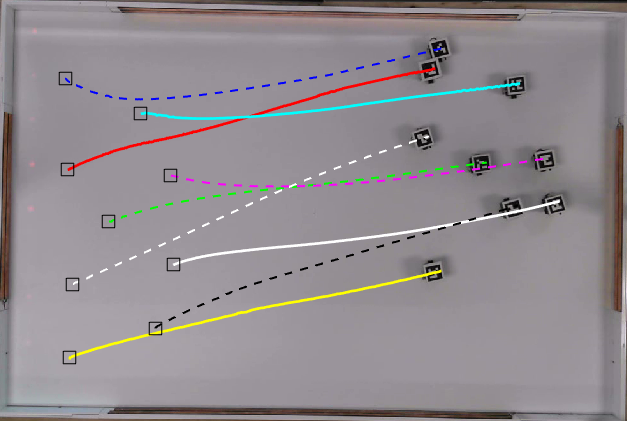} \label{fig:external_users_attitude_synchronization}}
	\caption{Experimental data from external user experiments rendered onto images of the Robotarium testbed setup. The square markers and curves are initial positions and trajectories of the GRITSBots, respectively. \label{fig:external_users}}
\end{figure*}

\subsubsection{Distributed Formation Control of Cyclic Formations from the University of Texas, Dallas} 
This experiment instantiated a distributed formation control algorithm for regular polygonal formations, originally presented in \cite{Fathian2016,Fathian2016a}. The controller uses relative position measurements in local coordinate frames and prohibits any inter-agent communication. 
Note that \cite{Fathian2016a} assumes agents to be points in the plane and does not consider collision avoidance. The successful execution on the Robotarium therefore depended on the use of Robotarium-provided barrier certificates shown in Section \ref{subsec:safety_barrier_certificates} and a mapping from single-integrator dynamics to the unicycle dynamics of the robots. Figure \ref{fig:external_users_formation_control} shows the results of this experiment with six robots.

\begin{table}[tbp]
    \centering
    \begin{tabular}{R{0.9cm} R{2.9cm} R{3.1cm}}
    \hline
    \bfseries Swarm Size $N$ & \bfseries {\centering Centralized Certificates $T_c$ (ms)} &  \bfseries Decentralized Certificates $T_d/N$ (ms) \\
    \hline
    10 	& 5.6   	& 3.2 \\
    \rowcolor{LightGray}
    40 	& 11.6		& 3.5 \\
    100 & 78.0		& 5.4 \\
  \end{tabular}
  \caption{Computation time of barrier certificates per iteration.}
  \label{table:qptime}
\end{table}

\subsubsection{Fault-tolerant Rendezvous from the University of Illinois Urbana-Champaign}
A second experiment instantiated a fault-tolerant version of the rendezvous algorithm, originally presented in \cite{Park2016}. In this work, agents achieve consensus by moving towards points within a safe set, while maintaining connectivity through extendable sensing capabilities \cite{Park2016}. 
Because this algorithm models agents as points in the plane and contains no native collision avoidance, the successful execution utilize the single-integrator-to-unicycle mapping and barrier certificates provided by the Robotarium. Figure \ref{fig:external_users_consensus} shows the results of this experiment with six robots.

\subsubsection{Passivity-based Attitude Synchronization from the Tokyo Institute of Technology}
A passivity-based attitude synchronization algorithm, originally presented in \cite{Igarashi2009}, was implemented on the Robotarium. Utilizing the passivity property of general rigid-body motion in $SE(3)$, this algorithm was designed to achieve attitude synchronization for a group of rigid bodies with only local information exchanges, i.e., $v_i=v_j,\lim_{t\to\infty}\theta_i(t)-\theta_j(t)=0,\forall i\neq j$.
The successfully execution of this algorithm relied on the following capabilities provided by the Robotarium: 1) specification of a local information exchange graph, i.e., a cycle graph ($C_N$); 2) a mapping from single-integrator to unicyle dynamics. Figure \ref{fig:external_users_attitude_synchronization} shows the results of this experiment with eight robots.

\subsection{The Simulation-Hardware Gap}
The Robotarium's infrastructure (see Section \ref{subsec:prototype_design}) is set up such that users prototype their code using the provided simulators and submit the exact same code for execution on the Robotarium's hardware. As such, it is important that the simulators provide a reasonably accurate approximation of the robots behavior. To characterize this simulation-hardware gap, we use linear regression on recorded data to provide a measure of system identification. This characterization relies on the GRITSBots dynamical model, i.e. the unicycle model given by
\begin{equation}
	\dot{x} = 
    \begin{bmatrix}
    	\dot{x}_{1} \\ 
        \dot{x}_{2} \\ 
        \dot{x}_{3}
    \end{bmatrix} = 
    \begin{bmatrix}
    	\cos(\theta) & 0 \\ 
        \sin(\theta) & 0 \\ 
        0 & 1
    \end{bmatrix}
    \begin{bmatrix}
    	v \\ 
        \omega 
    \end{bmatrix},
    \label{eq:unicycle-model}
\end{equation}
where $\theta$ is the orientation of the robot, $v$ and $\omega$ are its linear and angular velocity, respectively. We assume that each linear regression coefficient appears in the observation model as $\hat{\dot{x}}_{i} = \alpha_{i}\dot{x}_{i}$, where $\hat{\dot{x}}_{i}$ is an observation of $\dot{x}_{i}$ that is obtained through the Euler method applied to observed position and orientation values. 
Let $\hat{\dot{X}}_{i} \in \mathbb{R}^{d},~i \in \{1,2,3\}$ be the collection of observations $\hat{\dot{x}}_{i}$ and $\dot{X}_{i} \in \mathbb{R}^{d}$ be the collection of evaluations of the unicycle model $\dot{x}_{i}$, where $d$ is the number of data points (here, we use $d = 30000$ data points). Then each coefficient is determined by the least squares linear regression equation 
\begin{equation}
	\alpha_{i} = (\dot{X}_{i}^{T}\dot{X}_{i})^{-1}\dot{X}_{i}^{T}\hat{\dot{X}}_{i},
    \label{eq:linear-regression}
\end{equation}
which yields the following modified unicycle model given by 
\[
	\dot{\bar{x}} = \mathrm{diag}(\alpha_1, \alpha_2, \alpha_3) \dot{x},
\]
where $\alpha_{1} = 0.8645, \alpha_{2} = 0.8119, \alpha_{3} = 0.4640$. The used data displayed a linear relationship between $\hat{\dot{x}}$ and $\dot{x}$, ensuring the accuracy of the linear regression.
The Robotarium simulators use these values for $\alpha_i$ together with Eqn. \eqref{eq:unicycle-model} to simulate the robot's dynamics. To evaluate the accuracy of the linear regression results, we furthermore implemented a waypoint-following algorithm (see \cite{Aicardi1995}). A video of this experiment is referenced in Table \ref{table:multimedia_extensions} while the following provides a numeric estimate of the error between simulation and hardware execution.
In particular, let $x_{sim}(t)$ be the trajectory in simulation and $x(t)$ the trajectory of the GRITSBot. Then we calculate the average error between the simulated and experimental trajectories as 
\begin{align}
	E(x_{sim}(t), x(t)) & = & \dfrac{1}{T}\int_{0}^{T} \min_{\tau \in [0, T]}\left(\|x_{sim}(\tau) - x(t)\|\right) dt,
    \label{eq:sim-exp-error}
\end{align}
which for this experiment yielded an error value of 
\[ 
E(x_{sim}(t), x(t)) = 0.0052~m,
\]
or an average difference of 5~mm between simulation and hardware execution.

\section{Conclusion}
\label{sec:conclusion}
In this paper, we have detailed the development of a remotely accessible, multi-robot research facility -- the \textit{Robotarium}. Beyond introducing the hardware and software components required to enable remote access of robot swarms, the Robotarium addressed the two key concerns of flexibility and safety. Unlike other remotely accessible testbeds, the Robotarium makes use of formal methods to ensure the safety of its physical assets and the avoidance of damage to the robots. These methods guarantee collision avoidance in a minimally invasive manner without overly constraining the type of control algorithms that can be executed on the Robotarium. To demonstrate the flexibility and versatility of this testbed as well as its use by the community, we have shown a number of external user examples that were deployed on the Robotarium with little implementation overhead and provable collision avoidance.

\begin{table}[tbp]
  \centering
  \begin{tabular}{ll}
    \hline
    \bfseries \bfseries Algorithm & \bfseries Link to Video \\
    \hline
    Safety barriers 			& \url{https://youtu.be/PWJk-oMcgn4} \\
    \rowcolor{LightGray}
    Long-term Operation 		& \url{https://youtu.be/PBdrD7ZS-qM} \\
    Track-following 			& \url{https://youtu.be/VIirTkWppkE} \\
    \rowcolor{LightGray}
    Dist. Formation Control		& \url{https://youtu.be/DXcF0h8Vld0} \\
    Fault-tolerant Consensus		& \url{https://youtu.be/AlUyUVoVMu0} \\
    \rowcolor{LightGray}
    Attitude Synchronization		& \url{https://youtu.be/cItg_vGv3jo} \\
  \end{tabular}
  \caption{List of video references.}
  \label{table:multimedia_extensions}
\end{table}

\section*{Acknowledgements}




We would like to thank Kaveh Fathian, Nicholas Gans, and Mark Spong from the University of Texas in Dallas, Hyongju Park and Seth Hutchinson from the University of Illinois in Urbana-Champaign and Junya Yamauchi and Masayuki Fujita from the Tokyo Institute of Technology for their participation in the early release of the Robotarium.

\bibliographystyle{IEEEtran}
\bibliography{references}

\end{document}